\documentclass{article}

\usepackage[preprint]{unireps_2023}

\usepackage[utf8]{inputenc} 
\usepackage[T1]{fontenc}    
\usepackage[backref=page]{hyperref}       
\usepackage{url}            
\usepackage{booktabs}       
\usepackage{amsfonts}   
\usepackage{amsmath}
\usepackage{nicefrac}       
\usepackage{microtype}      
\usepackage{xcolor}         
\usepackage{graphicx}
\usepackage{subfig}
\usepackage[normalem]{ulem}
\usepackage{wrapfig}
\usepackage{caption}
\usepackage{capt-of}
\usepackage{xfrac}
\usepackage{tabularray}
\usepackage{enumitem}

\title{On consequences of finetuning on data with highly discriminative features}

\begin{document}

\author{%
  Wojciech Masarczyk\textsuperscript{1,*}
  \And
  Tomasz Trzciński\textsuperscript{1,4,6}
  \And
  Mateusz Ostaszewski\textsuperscript{1}
}

\footnotetext[1]{\small Warsaw University of Technology, Poland}
\footnotetext[2]{\small IDEAS NCBR, Poland}
\footnotetext[3]{\small Tooploox, Poland}
\renewcommand*{\thefootnote}{\fnsymbol{footnote}}
\footnotetext[1]{\small Corresponding author: \texttt{wojciech.masarczyk@gmail.com}}

\maketitle


\section{Introduction}

Deep learning has witnessed remarkable advancements in various domains, driven by the ability of neural networks to learn intricate patterns from data. One key aspect contributing to their success is the process of transfer learning, where pre-trained models are fine-tuned on specific tasks, leveraging knowledge acquired from previous training~\cite{pratt1996transfer,yosinski2014transferable}. This technique is especially important in the advent of training ever-growing models such as Large Language Models (LLMs)~\cite{devlin2019bert,anil2023palm,openai2023gpt4} or massive ViTs~\cite{zhai2022scaling, dehghani2023scaling}. However, while transfer learning is a powerful tool, it is not without its nuances.

\begin{wrapfigure}[20]{r}{0.55\textwidth}
    \centering
  \includegraphics[width=0.55\textwidth]{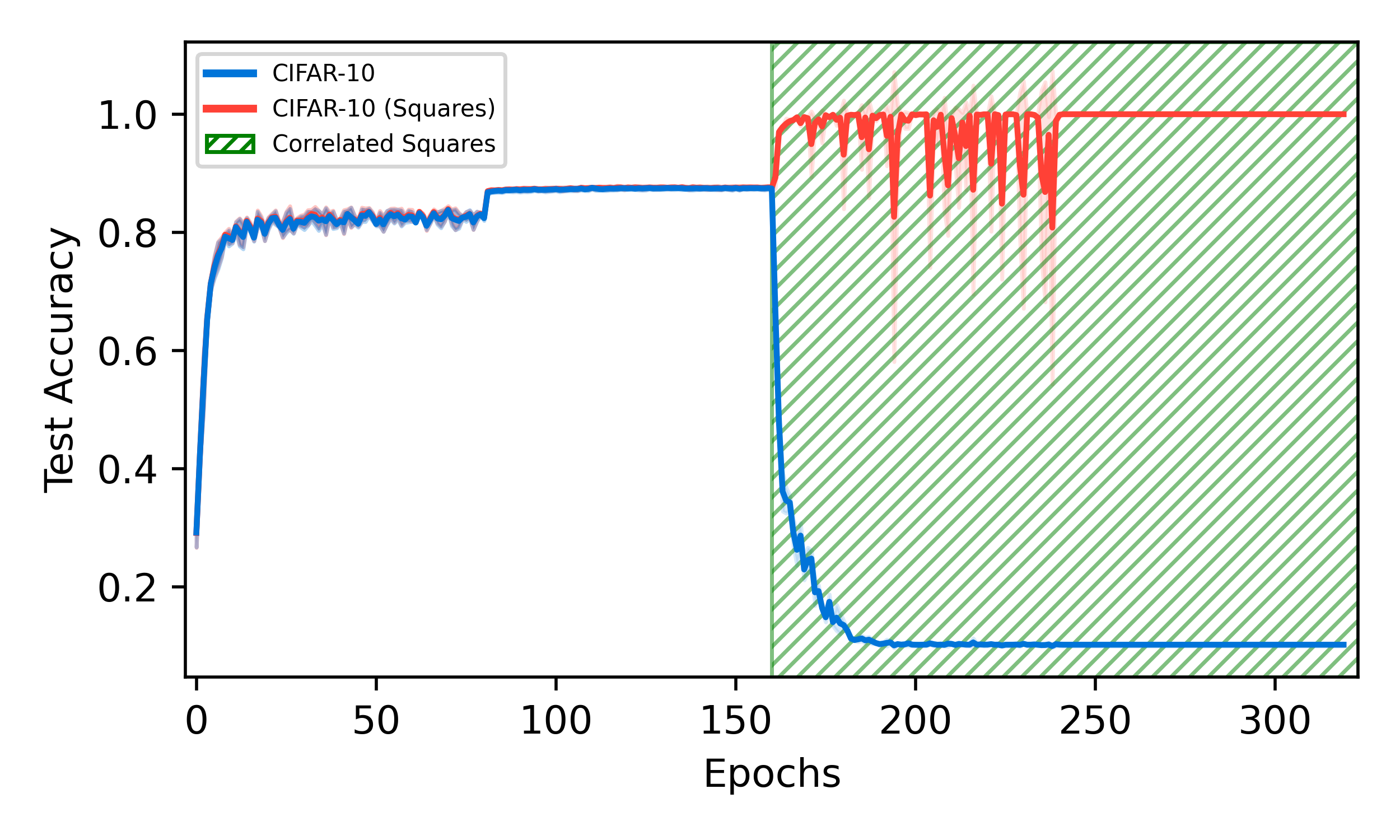}
  
  \caption{ \textbf{Feature erosion for VGG-19 on CIFAR-10}. Introducing a simpler discriminative feature (green hatch area) solely focuses the model on that feature (red curve) and abruptly erases previous knowledge about the data (blue line). \label{fig:teaser}}
\end{wrapfigure}

This work presents a thought-provoking experiment exposing a network's tendency to greedily follow the simplest discriminative features present in the data. This phenomenon was observed in multiple works and is commonly called simplicity bias~\cite{arpit2017closer,vallepérez2019deep,nakkiran2019sgd}. Here, we take a step further and investigate the implications of this behavior in the realms of transfer learning. 

To investigate this effect, we train the network on CIFAR-10 to perfect training accuracy (error-free).
Next, we introduce a highly discriminative, class-correlated pattern to the corner of each dataset image and proceed with the training. Surprisingly, as shown in Fig.~\ref{fig:teaser}, despite the model's perfect accuracy, finetuning it on the oversimplified task causes an abrupt performance loss (blue curve) and pushes the model to focus solely on the novel pattern. We call this phenomenon \textit{feature erosion}.
Our analysis shows that during the fine-tuning phase, the pretrained model greedily abandons salient, generalizing features in favor of the new discriminative ones.



\begin{figure}[!h]
    \centering
    \subfloat[ResNet 18 - CIFAR-10]
    {
      \includegraphics[width=0.32\textwidth]{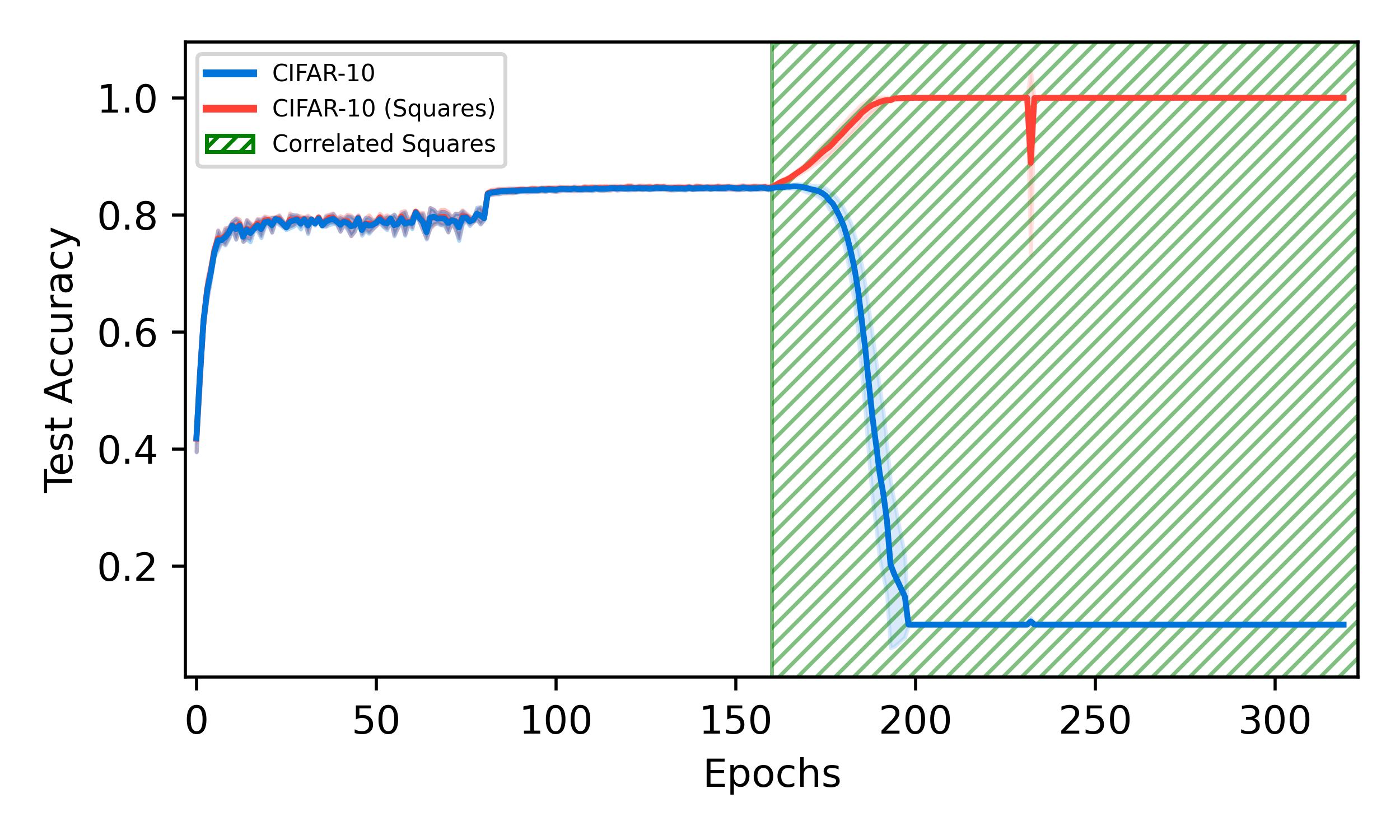}
    }
    \subfloat[ResNet 50 - ImageNet]
    {
      \includegraphics[width=0.32\textwidth]{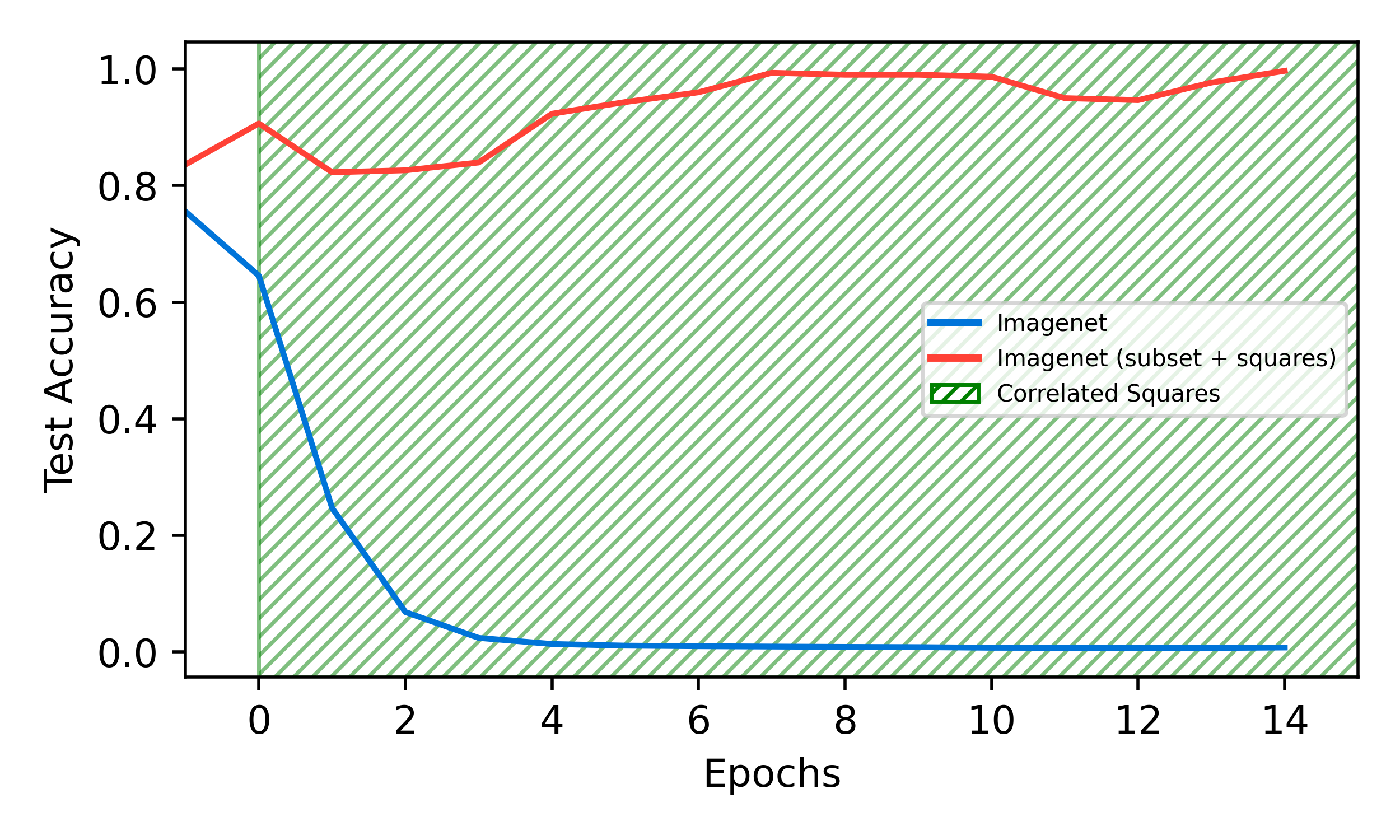}
    }
    \subfloat[VGG-19 - ImageNet]
    {
      \includegraphics[width=0.32\textwidth]{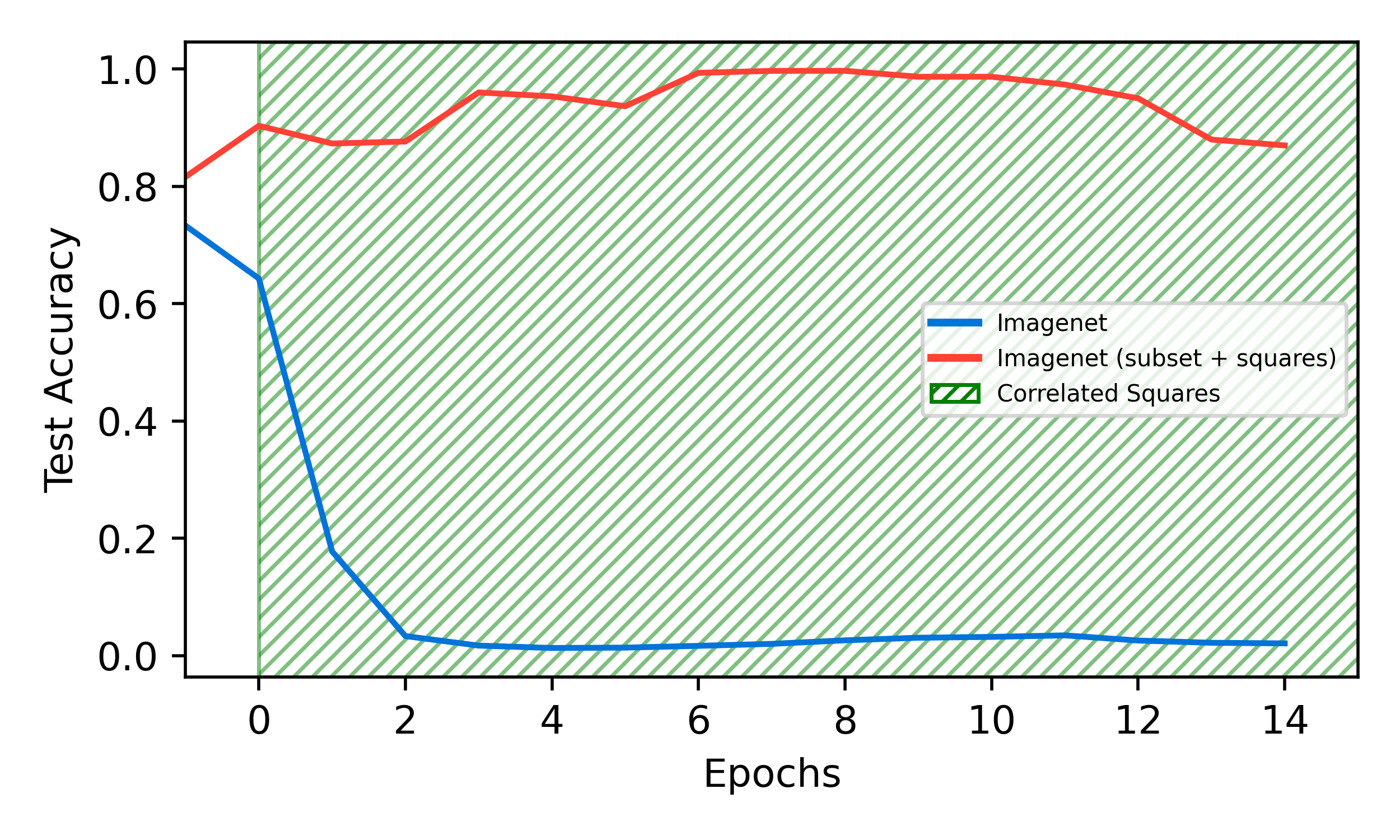}
    }
    \caption{\looseness-1 \small Feature erosion effect for ResNet18 trained on CIFAR-10 (left) ResNet50 trained of ImageNet (middle) and VGG-19 trained on ImageNet (right). ImageNet models used pretrained weights provided by PyTorch. ResNet18 reached 100\% training accuracy before introducing the discriminative pattern to the dataset.   } 
    \label{fig:datasets_architectures}
\end{figure}

In Section~\ref{sec:experiments}, we define details of the experiments and investigate the breadth of the phenomenon~\ref{fig:datasets_architectures}, presenting that all tested contemporary neural networks exhibit this behavior. Next, we investigate the detrimental effect of feature erosion on the model's representation formation, transfer learning, and plasticity. Section~\ref{sec:discussion} discusses the phenomenon's implications, its novelty concerning related works, and its hazards to real-world applications. 




\newpage
\section{Experiments and results}\label{sec:experiments}

\begin{wrapfigure}[16]{r}{0.4\textwidth}
    \vspace{-1.5cm}
    \centering
  \includegraphics[width=0.4\textwidth]{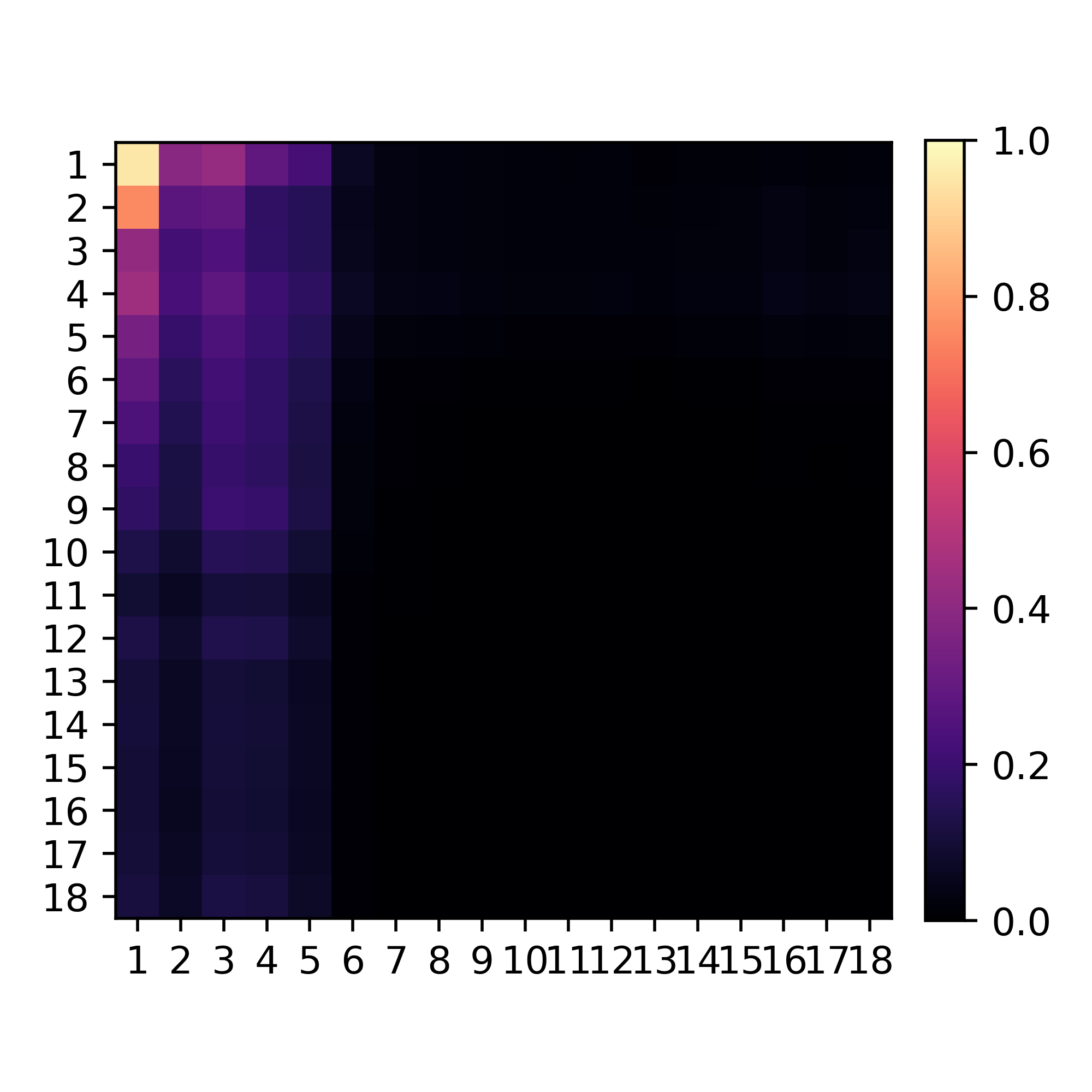}
  
  \vspace{-0.7cm}
  \caption{\looseness-3 \small \textbf{Feature erosion impacts CKA similarity between representations} from different layers extracted from the model after completing the 1st and 2nd tasks. The experiment was conducted using the VGG-19 model and the CIFAR-10 dataset. \label{fig:cka}}
\end{wrapfigure}

\begin{wrapfigure}[17]{r}{0.55\textwidth}
    \centering
    \includegraphics[width=0.49\textwidth]{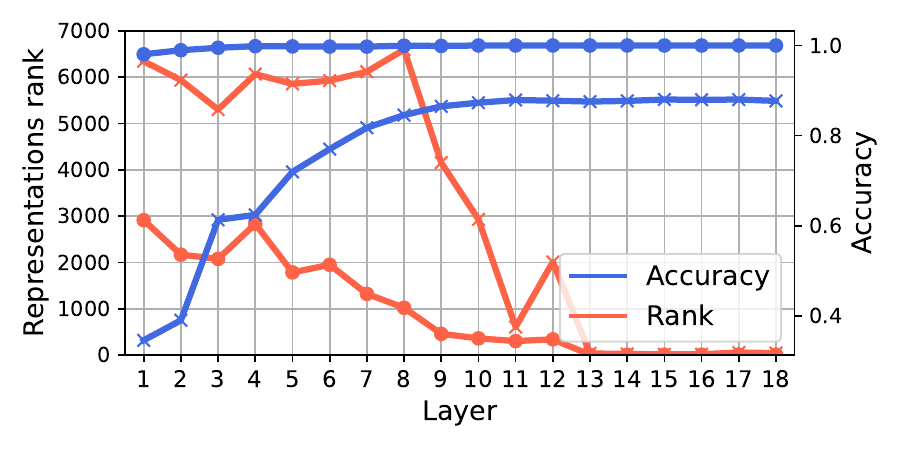}
    \caption{\looseness-3 \small \textbf{Feature Erosion collapses rank of the hidden representations} and impacts linear probing accuracy of VGG-19 trained on CIFAR10. Crosses refer to model after pre-training, dots refer to the model after finetuning. Blue color refer to the test accuracy, orange color refers to the representations rank. }
    \label{fig:id_and_rank}
\end{wrapfigure}

\begin{figure}[!h]
    \centering
  \includegraphics[width=0.55\textwidth]{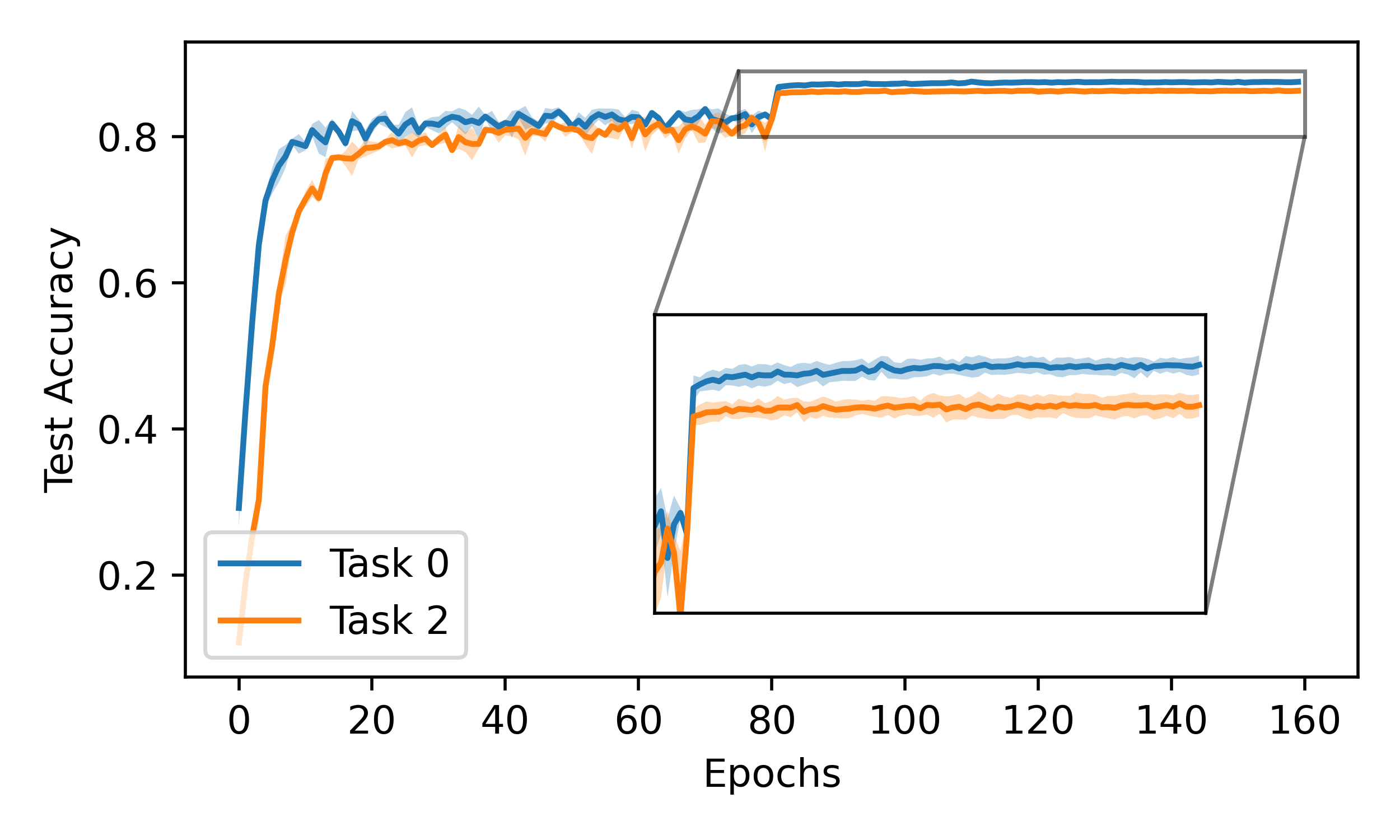}

  \caption{\textbf{Loss of plasticity}. The model is trained on the sequence of 3 tasks. The first (Task 0) and the last (Task 2) are standard CIFAR-10 datasets. The middle task is CIFAR-10 with correlated squares. \label{fig:plasticity} }
\end{figure}

In this section, we explore the phenomenon we introduced in the previous section, aiming to understand its severity and impact on network behavior. We initiate our investigation by examining the robustness of feature erosion across different network architectures and datasets (Fig.~\ref{fig:datasets_architectures}). Next, we assess the detrimental effect of the phenomenon on the network's representations (Fig.~\ref{fig:cka}) and show that pursuing simpler, discriminative patterns collapse the network's rank (Fig.~\ref{fig:id_and_rank}). We hypothesize that this effect is linked with the loss of plasticity observed in further training of that network (Fig.~\ref{fig:plasticity}).

\textbf{Experimental setup} We will now delve into feature erosion using ResNet-18, ResNet-50, and VGG-19 models trained on either the CIFAR-10 or ImageNet dataset. For CIFAR-10, these networks underwent 160 training epochs, consistently achieving $100\%$ training accuracy before introducing the oversimplified second task. The same hyperparameters were maintained for an additional 160 epochs during the second task, following recommendations for optimal model performance by Liu et al.~\cite{liu2018rethinking}.

Regarding the ImageNet models, we utilized PyTorch's pre-trained weights and randomly selected a subset of 10 classes from the ImageNet dataset for the oversimplified task. To create this dataset, we superimposed squares of the same size and placement on the training images, with each square's color representing the image's class. In most of our experiments, we applied these squares to all training images, the exception is presented in Fig.~\ref{fig:percent_of_squares}, where we investigated the impact of that ratio on the model's performance.





\textbf{Feature Erosion for different models and datasets}

The performance of these models is illustrated in Figure~\ref{fig:datasets_architectures}, showcasing test accuracy curves during both the pretraining phase (white background) and the subsequent fine-tuning phase on the oversimplified task (green-hatched background). The results clearly indicate that all neural architectures and datasets exhibit feature erosion, resulting in a noticeable decline in test accuracy on the pre-training dataset. Notably, as the sole distinction between the first and second tasks is the presence of these colored squares, the dramatic shift from near-perfect accuracy in the first task to random-chance accuracy in the second task implies that the model exclusively focuses on the colored squares.

\textbf{Unpacking Feature Erosion: Analyzing Representations}

Having observed significant shifts in model performance on established benchmarks, our objective is to investigate the impact of fine-tuning on oversimplified datasets on the model's representations.

In our subsequent experiment, we perform a comparative analysis of representations at each layer after training on the first task and subsequent training on the second task. We employ Centered Kernel Alignment (CKA)~\cite{kornblith2019similarity} as a metric to quantify similarity. We extract representations from standard CIFAR-10 datasets (without color squares) for both models to isolate the impact of weight evolution on model representations.

As illustrated in Figure~\ref{fig:cka}, the representations show substantial dissimilarity. Most notably, most diagonal elements are black, indicating minimal similarity between representations within the same layers after undertaking different tasks. The lighter colors in the top left corner suggest that changes during the fine-tuning phase commence early in the network, particularly in the bottom layers.

Given the observed dramatic changes in representations across nearly all layers, we delve deeper into understanding the underlying dynamics. In Fig.~\ref{fig:id_and_rank}, we investigate feature erosion within each layer using linear probing and the numerical rank of representations. A comparison of the linear probing plots reveals that after fine-tuning (blue dots), the model fully adapts to the new data, with the second layer achieving accuracy levels comparable to those of the entire model. Furthermore, the numerical rank of representations experiences a substantial decline at each layer following fine-tuning (red dots), indicating that the model, starting from the initial layers, probably adheres to the simplicity bias~\cite{morwani2023simplicity} and projects the data into smaller subspaces.

\textbf{Loss of plasticity}

To better understand the impact of feature erosion on network performance, we examined the network's ability to relearn information from a previous task. We conducted experiments involving a sequence of three tasks: CIFAR-10 (Task 0), an oversimplified version of CIFAR-10, and the standard CIFAR-10 once again (Task 2).

Typically, relearning is expected to be faster and require fewer computational resources than training from scratch. However, our results, as shown in Figure 1, indicate a deviation from this expected behavior. In this setup, the network not only learns more slowly compared to training from scratch but also fails to achieve the same level of performance within the same computational budget as the network trained from scratch. This performance difference is commonly referred to as "loss of plasticity" and is often associated with the degradation of the penultimate layer's rank. While we speculate that this explanation may apply in our case, a thorough investigation of this hypothesis is beyond the scope of our current research.


\section{Discussion}\label{sec:discussion}



In this study, we have uncovered a nuanced aspect of catastrophic forgetting, demonstrating that it can occur even when the full dataset is present in the new task data. Our experiments, which include GradCam analysis, reveal that despite achieving 100\% training accuracy after introducing highly discriminative features to the model, it begins to focus exclusively on these new features, leaving behind previously developed ones. A deeper investigation using techniques such as Centered Kernel Alignment (CKA) exposes significant changes in representations, along with a deterioration in model performance indicated by a rank collapse of representations in nearly all layers.



Our research expands our knowledge of the complex relationship between
task similarity~\citep{ramasesh2020anatomy,braun2022exact}, forgetting, and transfer dynamic~\citep{chen2022forgetting}. On the one hand, recent studies have revealed that intermediate task similarity tends to contribute most to catastrophic forgetting~\citep{ramasesh2020anatomy,braun2022exact}. The task similarity is the use of a "data-mixing framework," which combines images and labels from two distinct datasets of equal size. However, our experimental setup features identical datasets, differing only in a small image segment transitioning from random to highly correlated with a specific class. While this does not contradict earlier findings, it certainly introduces a novel perspective on the phenomenon.
On the other hand, our observation also has implications in light of recent research~\citep{chen2022forgetting}, which suggests that less forgetful representations result in improved performance on new tasks, indicating a robust relationship between retaining previous information and enhanced learning efficiency. In this context, our toy example falsifies the reverse implication, i.e., the model exhibits perfectly transferable features yet forgets them in favor of features with greater predictive power.

\begin{wrapfigure}[19]{r}{0.45\textwidth}
    \centering
    \vspace{-0.5cm}
    \includegraphics[width=0.45\textwidth]{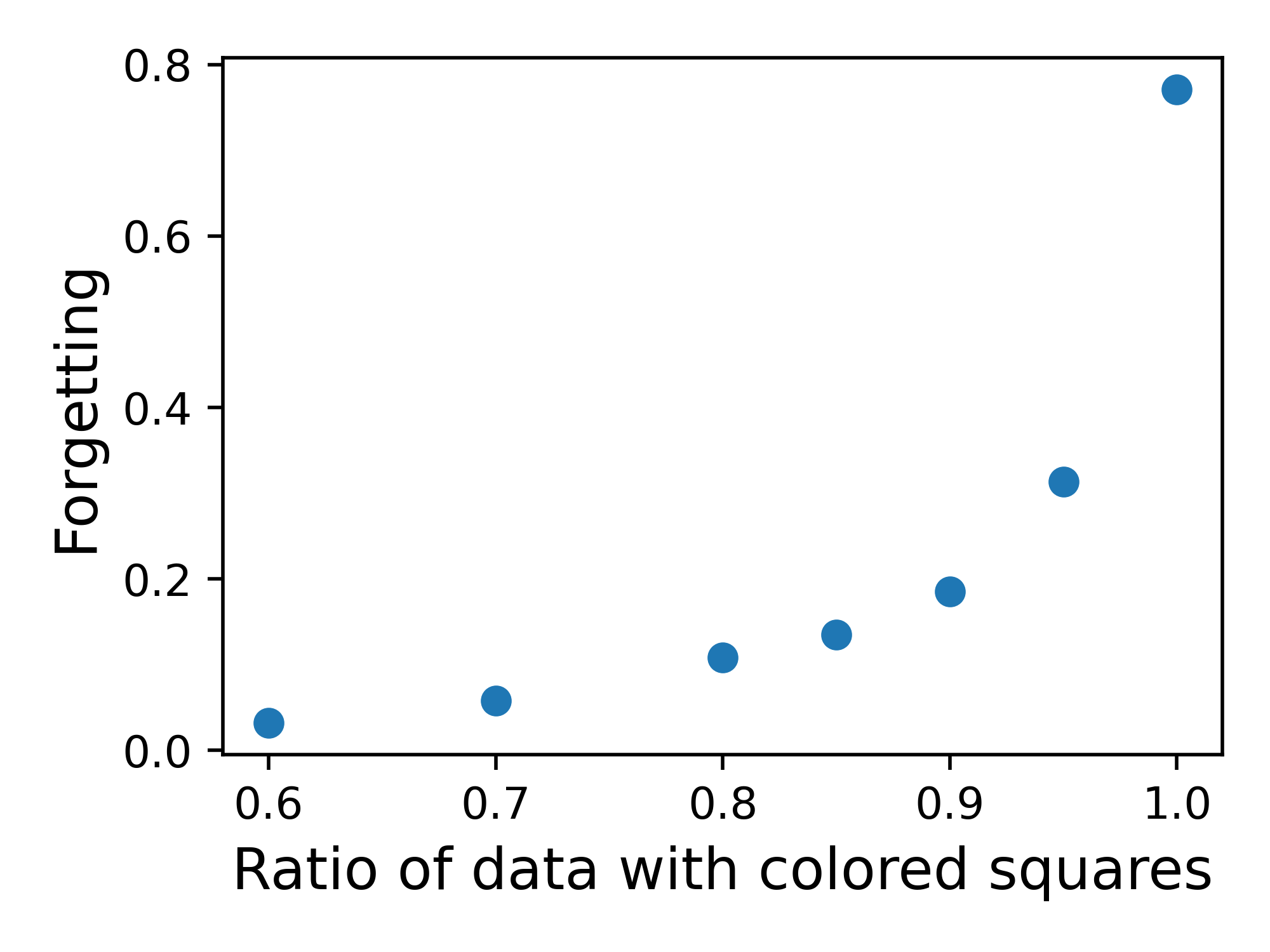}
    \caption{The stronger the discriminative pattern (higher ratio), the higher the forgetting of the model. Each dot represents a single model trained on CIFAR-10 and fine-tuned on an oversimplified task with different ratios of images with colored squares.}
    \label{fig:percent_of_squares}
\end{wrapfigure}

In our concluding experiment, extending our analysis of feature erosion, we explore the relation between forgetting and the ratio of samples containing oversimplified discriminative patterns. In Fig.~\ref{fig:percent_of_squares}, we observe a non-linear relationship between the number of oversimplified samples in the training dataset and the extent of forgetting. However, even modest ratios of oversimplified data are enough to induce forgetting in the model. This reinforces the importance of the phenomenon. 

The situation we present in this study has practical significance, especially in scenarios involving incremental learning in various domains. Real-world applications that continually receive new data with limited human involvement might come across samples containing highly predictive patterns. This can lead to the loss of previously acquired knowledge. For instance, in medical imaging, where data artifacts may correlate with task objectives, the phenomenon of feature erosion could be a frequent concern.

Additionally, our findings connect with existing literature on concepts such as simplicity bias~\citep{neyshabur2014search,shah2020pitfalls} and gradient starvation~\citep{pezeshki2021gradient}. Our results suggest that simplicity bias not only affects generalization but can also disrupt previously well-functioning representations. The resilience of simplicity bias to approaches like ensembles or adversarial training raises questions about the effectiveness of common continual learning methods.

Finally, there may be a positive aspect to this phenomenon. In the current era of heightened focus on AI ethics, machine unlearning~\citep{bourtoule2021machine} and fairness in deep learning~\citep{du2020fairness} are prominent topics. Our study prompts the question of whether intentionally introducing highly discriminatory patterns to unwanted samples can facilitate the intentional forgetting of such samples, a topic that warrants further exploration.

In summary, our work reveals a novel facet of catastrophic forgetting, challenging conventional wisdom about its occurrence and implications. These findings have relevance for both the field of machine learning and practical applications that involve continual learning with evolving data.


\bibliography{bibliography}
\bibliographystyle{plainnat}

\end{document}